\documentclass[10pt,conference,a4paper]{IEEEtran}
\pdfoutput=1
\usepackage{cite}
\usepackage{amsmath}
\usepackage{amssymb}
\usepackage{graphicx}

\begin{document}
%
\title{Beam Search for Learning a Deep Convolutional Neural Network of 3D Shapes }

\author{\IEEEauthorblockN{Xu Xu and Sinisa Todorovic}
\IEEEauthorblockA{School of Electrical Engineering and Computer Science\\
Oregon State University, Corvallis, Oregon 97330\\
Email: xuxu@oregonstate.edu, sinisa@eecs.oregonstate.edu}
}

\maketitle

\begin{abstract}
This paper addresses 3D shape recognition. Recent work typically represents a 3D shape as  a set of binary variables corresponding to 3D voxels of a uniform 3D grid centered on the shape, and resorts to deep convolutional neural networks (CNNs) for modeling these binary variables. Robust learning of such CNNs is currently limited by the small datasets of 3D shapes available -- an order of magnitude smaller than other common datasets in computer vision. Related work typically deals with the small training datasets  using a number of ad hoc, hand-tuning strategies. To address this issue, we formulate CNN learning as a beam search aimed at identifying an optimal CNN architecture -- namely, the number of layers, nodes, and their connectivity in the network -- as well as estimating parameters of such an optimal CNN. Each state of the beam search corresponds to a candidate CNN. Two types of actions are defined to add new convolutional filters or new convolutional layers to a parent CNN, and thus transition to children states. The utility function of each action is efficiently computed by transferring parameter values of the parent CNN to its children, thereby enabling an efficient beam search. Our experimental evaluation on the 3D ModelNet dataset demonstrates that our model pursuit using the beam search yields a CNN with superior performance on 3D shape classification than the state of the art.
\end{abstract}


%
\IEEEpeerreviewmaketitle

\section{Introduction} \label{sec1}

This paper addresses the problem of 3D shape classification. Our goal is to predict the object class of a given 3D shape, represented as a set of binary presence-absence indicators associated with 3D voxels of a uniform 3D grid centered on the shape. This is one of the basic problems in computer vision, as 3D shapes are important visual cues for image understanding. This is a challenging problem, because an object's shape may significantly vary due to changes in the object's pose and articulation, and appear quite similar to shapes of other objects.

There is a host of literature on reasoning about 3D object shapes \cite{Besl:1992, Tangelder:2008}. Traditional approaches typically extract feature points from a 3D shape \cite{wks,hks,curvature}, and find correspondences between these feature points for shape recognition and retrieval \cite{RF,spectral,fgm}. However, these methods tend to be sensitive to long-range non-rigid and non-isometric shape deformations, because, in part, the feature points capture only local shape properties. In addition, finding optimal feature correspondences is often formulated as an NP-hard non-convex Quadratic Assignment Problem (QAP), whose efficient solutions come at the price of compromised accuracy. 

Recently, the state-of-the-art performance in 3D shape classification has been achieved using deep 3D Convolutional Neural Networks (CNNs), called 3D ShapeNets \cite{3dshapenets}. 3D ShapeNet extends the well-known deep architecture called AlexNet \cite{krizhevsky2012imagenet}, which is widely used in image  classification. Specifically, the extension modifies 2D convolutions computed in AlexNet to 3D convolutions. However, 3D ShapeNets' architecture consists of 3 convolutional layers and 2 fully-connected layers, with a total of 12M parameters. This in turn means that a robust learning of ShapeNet parameters requires large training datasets. But in comparison with common datasets used in other domains, currently available 3D shape datasets are smaller by at least an order of magnitude. For example, the benchmark 3D shape datasets SHREC'14 dataset \cite{Li20151} has only 8K shapes, and ModelNet dataset \cite{3dshapenets} has 150K shapes, whereas the well-known ImageNet \cite{russakovsky2015imagenet} has 1.5M images.

Faced with small training datasets, existing deep-learning approaches to 3D shape classification typically resort to a number of ad hoc, hand-tuning strategies for robust learning. Rarely are these justified based on extensive empirical evaluation, as it would take a prohibitively long time, but on past findings of other related work (e.g., image classification). Hence, their particular design choices about the architecture and learning -- e.g.,  the number of convolutional and fully-connected layers used, or specification of the learning rate in backpropagation -- may be suboptimal.

Motivated by the state-of-the-art performance of 3D ShapeNets \cite{3dshapenets}, we here adopt this framework, and focus on addressing the aforementioned issues in a principled manner. Specifically, we formulate a model pursuit for robust learning of 3D CNN. This learning is specified as a beam search aimed at identifying an optimal CNN architecture -- namely, the number of layers, nodes, and their connectivity in the network -- as well as estimating parameters of such an optimal CNN. Each state of the beam search corresponds to a candidate CNN. Two types of actions are defined to add either new convolutional filters or new convolutional layer to a parent CNN, and thus transition to children states. The utility function of each action is efficiently estimated as a training accuracy of the resulting CNN. The efficiency is achieved by transferring parameter values of the parent CNN to its children. Starting from the root ``shallow and narrow'' CNN, our beam search is guided by the utility function toward generating more complex CNN candidates with an increasingly larger classification accuracy on 3D shape training data, until training accuracy stops increasing. The CNN candidate with the highest training accuracy  is finally taken as our 3D shape model, and used in testing.

In our experimental evaluation on the 3D ModelNet dataset \cite{3dshapenets}, our beam search yields a 3D CNN with $150$ times fewer parameters than 3D ShapeNets \cite{3dshapenets}. The results demonstrate that our 3D CNN outperforms 3D ShapeNets by 3\% on 40 shape classes. This suggests that a model pursuit using beam search is a viable alternative to currently heuristic practice in designing deep CNNs.

In the following, Sec.~\ref{sec2} gives an overview of our 3D shape classification using 3D CNN; Sec.~\ref{sec3} formulates our beam search in terms of the state-space, successor function, heuristic function, lookahead and backtrack strategy; Sec. \ref{sec4} specifies our efficient transfer of parameters from a parent model to its children in the beam search; and Sec. \ref{sec5} presents our results.

\section{3D Shape Classification Using 3D CNN} \label{sec2}

For 3D shape classification we use a 3D CNN. Given a binary volumetric representation of a 3D shape as input, our CNN predicts an object class of the shape. Below, we first describe the shape representation, and then explain the architecture of 3D CNN.

In this paper, we adopt the binary volumetric representation of 3D shapes presented in \cite{3dshapenets}.  Specifically, each shape is represented as a set of binary indicators corresponding to 3D voxels of a uniform 3D grid centered on the shape. The indicators take value 1 if the corresponding 3D voxels are occupied by the 3D shape; and 0, otherwise. Hence, each 3D shape is represented by a binary three-dimensional tensor. The grid size is set to $30 \times 30 \times 30$ voxels. The shape size is normalized such that a cube of $24 \times 24 \times 24$ voxels fully contains the shape, and the remaining empty voxels serve for padding in all directions around the shape. Each shape is also labeled with a corresponding object class.

As mentioned in Sec.~\ref{sec1}, the architecture of our 3D CNN is similar to that of 3D ShapeNets \cite{3dshapenets}, with the important distinction that we greatly reduce the total number of parameters. As we will discuss in greater detail in the results section, the beam search that we use for the model pursuit yields a 3D CNN with 3 convolutional layers and 1 fully connected layer, totaling 80K parameters. The top layer of our model represents the standard soft-max layer for classifying the input shapes into one of possible object classes.

In the following section, we specify our model pursuit and the initial root CNN from which the beam search starts exploring candidate, more complex models, until training error cannot be further reduced.

\section{Beam Search}\label{sec3}

Search-based approaches have a long-track record of successfully solving computer vision problems, including structured prediction for scene labeling\cite{Lam_2015_CVPR, Roy_2014_CVPR}, object localization \cite{lampert2009efficient}, and boundary detection \cite{payet2013sledge}. Unlike the above related work, search in this paper is not used for inference, but for identifying an optimal CNN architecture and estimating CNN parameters. For efficiency of learning, we consider a beam search which limits the exploration of the state space to a few top candidates. Our beam search is defined by the following: 
\begin{itemize}
\item States correspond to CNN candidates,
\item Initial state represents a small CNN,
\item Successor function generates new states based on actions taken in parent states,
\item Heuristic function evaluates the utility of the actions, and thus guides the beam search,
\item Lookahead and backtracking strategy.
\end{itemize}

\begin{figure}[ht]
\centering
\includegraphics[width=0.9\linewidth]{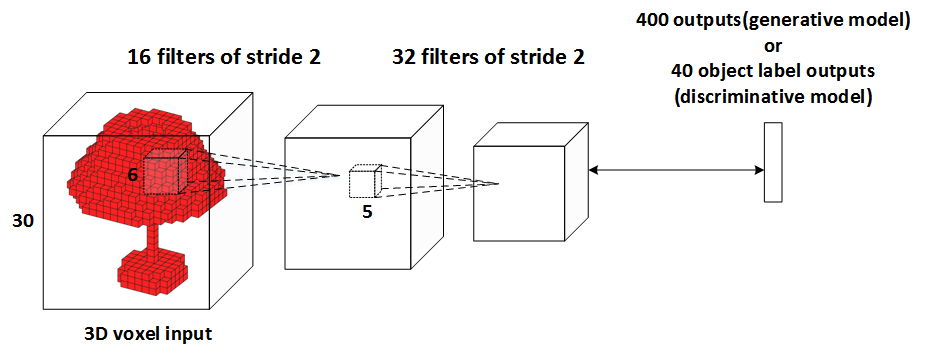}
\caption{Architecture of our initial CNN model.}
\label{fig:arch}
\end{figure}

\textbf{State-space:} The state-space is defined as $\Omega = \{\textbf{s}\}$, where state $\textbf{s}$ represents a network configuration (also called architecture). A CNN's network configuration specifies the number of convolutional and fully-connected layers, the number of hidden units or 3D convolutional filters used in each layer, and which layers have max-pooling. In this paper, we constrain the beam search such that the size of the fully connected layer remains the same as in the initial CNN, because we have empirically found that only extending convolutional layers maximally increases   the network's classification accuracy (as also reported in \cite{residual,vgg}).

\textbf{Initial State:}  Our model pursuit starts from a relatively simple initial CNN, illustrated in Figure \ref{fig:arch}, and the goal of the beam search is to extend the initial model by adding either new convolutional filters to existing layers, or new convolutional layers. The initial model consists of only two convolutional layers and one fully-connected layer. The first convolutional  layer has 16 filters of size 6 and stride 2. The second convolutional  layer has 32 filters of size 5 and stride 2. Finally, the fully-connected layer has 400 hidden units. 

The parameters of the initial CNN are trained as follows. We first generatively pre-train the model in a layer-wise fashion, and then use a discriminative fine-tuning procedure. The standard Contrastive Divergence \cite{hinton2002training} is used  to pre-train the two convolutional layers, whereas the top fully-connected layer is trained using Fast Persistent Contrastive Divergence \cite{tieleman2009using}. Once one layer is learned, the weights are fixed and the hidden activations are fed into the next layer as input. After this pre-training, we continue to discriminatively fine-tune the pre-trained model.   We first replace the topmost layer with a new randomly initialized fully-connected layer, and then add a standard softmax layer on top of the network to output class probabilities. The standard cross-entropy loss is computed using ground-truth class labels, and used in backpropagation to update the weights in all layers.

Given this simple, initial CNN, the beam gradually builds a search tree with new states \textbf{s}.  Exploration of the state space consists of generating successor states from a few selected parent states. The selection is based on ranking the parent states by a heuristic function, as further explained below.

\textbf{Successor function:} $\Gamma:\textbf{s} \rightarrow \textbf{s}'$, generates new states $\textbf{s}'$ from $\textbf{s}$ by applying an action $\textbf{a}\in \textbf{A}$ from a set of possible actions $\textbf{A}$. In this paper, we specify $\textbf{A}$ as consisting of two types of actions: 1) Add a new convolutional layer at the top of all existing convolutional layers, where the newly added layer has the same number of filters, filter size, and stride as the top convolutional layer; 2) Double the number of filters in the top convolutional layer. Other alternative definitions of $\textbf{A}$ are also possible. In particular, we have also considered an action which adds max-pooling to convolutional layers; however, such an extended $\textbf{A}$ has not produced better performance on test data, relative to the above case when only two types of actions are considered.

As one of our technical contributions, in this paper, we specify an efficient successor function for enabling an efficient beam search. Specifically, we apply a knowledge transfer procedure, following the approach of \cite{net2net}, which efficiently copies parameter values of the previous state $\textbf{s}$ to values of newly added parameters in the generated state $\textbf{s}'$. After this knowledge transfer, the new CNN $\textbf{s}'$ is fine-tuned using only a few iterations (in our experiments, the number of iterations is 10), for robustness. Note that a significantly larger number of iterations would have been necessary for this fine-tuning, had we randomly initialized the newly added parameters of $\textbf{s}'$ (as is common in practice), instead of using knowledge transfer. In this way, we achieve efficiency. In the following section, we explain our knowledge transfer procedure in more detail.

\textbf{Heuristic function:} $\mathcal{H}(\textbf{s},\textbf{s}')$ ranks new states $\textbf{s}'$ given their parent states $\textbf{s}$.  $\mathcal{H}(\textbf{s},\textbf{s}')$  is used to guide the beam search, which selects the top $K$ successor states, where $K$ is taken as a beam width. $\mathcal{H}(\textbf{s},\textbf{s}')$ is defined as the difference in classification accuracy on training data between $\textbf{s}$ and $\textbf{s}'$.

\textbf{Lookahead and backtracking strategy:} For robustness, we specify a lookahead and backtracking strategy for selecting the top $K$ successor states. We first explore the state space by applying the successor function several times from parent states $\textbf{s}$, until the resulting tree search reaches a depth limit, $D$. Then, among the leaf states $\textbf{s}'$ at the tree depth $D$, we select the top $K$ leaves $\textbf{s}'$ evaluated with $\mathcal{H}(\textbf{s},\textbf{s}')$. From these top $K$ leaf states, we backtrack to the unique $K$ children of parent states $\textbf{s}$, taken as valid new candidate CNNs.

\section{Knowledge Transfer} \label{sec4}

When generating new candidate CNNs, we make our beam search efficient by appropriately transferring parameter values from parent CNNs to their descendants. In the sequel, we specify this knowledge transfer for the two types of search actions considered in this paper.

\subsection{Net2WiderNet}
A new state can be generated by doubling the number of filters in the top convolutional layer of a parent CNN. This action effectively renders the new candidate CNN ``wider'' than its parent model, and hence we call this action {\em Net2WiderNet}. We estimate the parameters of the ``wider'' CNN as follows. 

The key idea is to estimate the newly added parameters such that the parent CNN and its ``wider'' child CNN give the same outputs for the same inputs. This knowledge-transfer strategy ensures that the newly generated model is not worse than the previously considered model. After this knowledge transfer, parameters of the ``wider'' child CNN $\textbf{s}'$ are fine-tuned to verify if the action resulted in a better model than the parent CNN $\textbf{s}$, as evaluated with the heuristic function $\mathcal{H}(\textbf{s},\textbf{s}')$.

In order to widen a convolutional layer, $i$, we need to update both sets of model parameters $\textbf{W}^{(i)}\in \mathbb{R}^{m\times n}$ and  $\textbf{W}^{(i+1)}\in \mathbb{R}^{n\times p}$ at layers $i$ and $i+1$, respectively, where layer $i$ has $m$ inputs and $n$ outputs, and layer $i+1$ has $p$ outputs. When the action {\em Net2WiderNet} extends layer $i$ so it has $q > n$ outputs, we define the random mapping function $g$ as 
\begin{equation}
    g(j)=
   \begin{cases}
   j &,\mbox{$0<j \leq n$}\\
   \text{random sample from } \{1,2, \cdots, n\} &,\mbox{$n < j \leq q$}
   \end{cases}
\end{equation}
Then, the new sets of parameters $\textbf{U}^{(i)}$ and $\textbf{U}^{(i+1)}$ can be computed from $\textbf{W}^{(i)}$ and $\textbf{W}^{(i+1)}$ as 
\begin{eqnarray}
\textbf{U}_{k,j}^{(i)} &=&  \textbf{W}_{k,g(j)}^{(i)}, \label{update1} \\
\textbf{U}_{j,h}^{(i+1)} &=& \dfrac{1}{|\{x|g(x)=g(j)\}|}  \textbf{W}_{g(j),h}^{(i+1)},  \label{update2}
\end{eqnarray}
where $k=1,...,m$, $j=1,...,q$, and $h=1,...,p$.

From (\ref{update1}), the first $n$ columns of $\textbf{W}^{(i)}$ are simply copied directly into $\textbf{U}^{(i)}$. Columns $n+1$ through $q$ of $\textbf{U}^{(i)}$ are created by randomly choosing columns of $\textbf{W}^{(i)}$, as defined in $g$. The random selection is performed with replacement, so each column of $\textbf{W}^{(i)}$ may be copied many times to columns $n+1$ through $q$ of $\textbf{U}^{(i)}$. 

From (\ref{update2}), we similarly have that first $n$ rows of $\textbf{W}^{(i+1)}$ are simply copied directly into $\textbf{U}^{(i+1)}$, and rows $n+1$ through $q$ of $\textbf{U}^{(i+1)}$ are created by randomly choosing rows of $\textbf{W}^{(i+1)}$, as defined in $g$. In addition, the new parameters in $\textbf{U}^{(i+1)}$ are normalized so as to account for the random replication of rows in $\textbf{U}^{(i+1)}$. The normalization is computed by diving the new parameters with a replication factor, given by  $\dfrac{1}{|\{x|g(x)=g(j)\}|}$. 

It is straightforward to prove that the resulting extended network with new parameters $\textbf{U}^{(i)}$ and $\textbf{U}^{(i+1)}$ produces the same outputs as the original network with parameters $\textbf{W}^{(i)}$ and $\textbf{W}^{(i+1)}$, for the same inputs.

\begin{figure}[ht]
\centering
\includegraphics[width=0.8\linewidth]{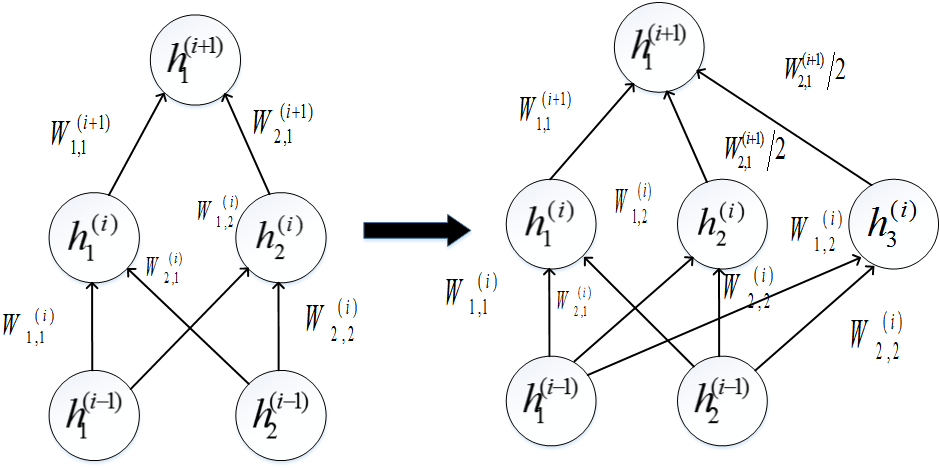}
\caption{An example of {\em Net2WiderNet} action.}.
\label{fig:net2widernet}
\end{figure}

An example of this procedure is illustrated by Figure \ref{fig:net2widernet}. In this example, we increase the size of hidden layer $h^{(i)}$ by adding one additional unit, while keeping the activations propagated to hidden layer $h^{(i+1)}$ unchanged. Assume that we randomly pick hidden unit $h_{2}^{(i)}$ to replicate, then we copy its weights $\textbf{W}_{1,2}^{(i)}$ and $\textbf{W}_{2,2}^{(i)}$ to the new $h_{3}^{(i)}$ unit. The weight $\textbf{W}_{2,1}^{(i+1)}$, going out of $h_{2}^{(i)}$, must be copied to also go out of  $h_{3}^{(i)}$. This outgoing weight must also be dived by 2 to compensate for the replication of  $h_{2}^{(i)}$.

\subsection{Net2DeeperNet}
The second type of action that we consider is termed {\em Net2DeeperNet}, since it adds a new convolutional layer to a parent CNN, thereby producing a deeper child CNN. Specifically, {\em Net2DeeperNet} replaces a layer $\textbf{h}^{(i)} = \phi(\textbf{W}^{(i)\top} \textbf{h}^{(i-1)})$ with two layers $\textbf{h}^{(i)} = \phi(\textbf{U}^{(i)\top}\phi(\textbf{W}^{(i)\top}\textbf{h}^{(i-1)}))$, where $\phi$ denotes the activation function. The new parameter matrix $\textbf{U}$ is specified as the identity matrix. 

\begin{figure}[ht]
\centering
\includegraphics[width=0.8\linewidth]{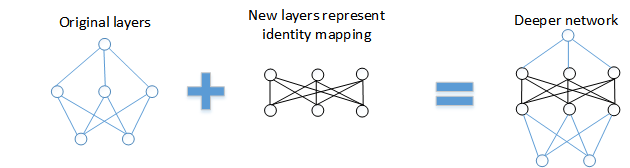}
\caption{An example of {\em Net2DeeperNet} action.}
\label{fig:net2deepernet}
\end{figure}

Figure \ref{fig:net2deepernet} shows an illustration of {\em Net2DeeperNet}. When we apply this action, we add a new convolutional layer and simply set the new convolution filters to be identity functions. A zero padding is also added to maintain the size of activations unchanged. 

It is worth noting that {\em Net2DeeperNet} does not guarantee that the resulting deeper network will give the same outputs as the original one, for the same inputs, when the activation function used is the sigmoid. The guarantee holds when the activation function used is the rectified linear unit (ReLU), though. However, in our experiments, we have not found that using the sigmoid hurts the specified knowledge transfer of {\em Net2DeeperNet} toward the efficient beam search.

\section{Experimental Results} \label{sec5}

\subsection{Dataset}
For evaluation, we use the ModelNet dataset \cite{3dshapenets}, and the same experimental set up of 3D ShapeNets \cite{3dshapenets}, for fair comparison.  ModelNet consists of 40 object classes such as chairs, tables,toilets, sofas, etc. Each class has 100 unique CAD models, representing the most common 3D shapes of the class, totaling 151,128 voxelized 3D models in the entire dataset. We conduct 3D classification on both the 10-class subset ModelNet10,  and the full 40-class dataset, as in  \cite{3dshapenets}. We use the provided voxelizations and train/test splits for evaluation. Specifically, for each class, 960 instances are used for training and 240 instances are used for testing. 

We have implemented our beam search in MATLAB, on top of a GPU-accelerated software library of 3D ShapeNets \cite{3dshapenets}. Experiments are run on a machine with the NVIDIA Tesla K80 GPU accelerator. 

\subsection{3D shape classification accuracy}

Our classification accuracy is averaged over all classes on test data, and used for comparison with 3D ShapeNets \cite{3dshapenets}. In addition, we average our classification accuracy over the five runs of the beam search from 5 different initial CNNs, all of which have the same architecture, but differently initialized parameters.

\begin{figure}[!h]
\centering
\includegraphics[width=0.7\linewidth]{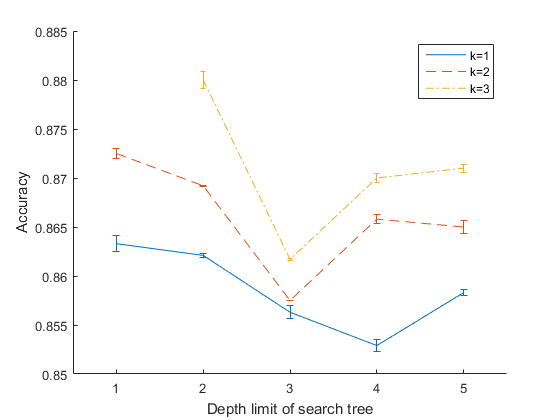}
\caption{3D shape classification accuracy on testing data of 10 classes at different beam width $\textit{K}$.}
\label{fig:test1}
\end{figure}

\begin{figure}[!h]
\centering
\includegraphics[width=0.7\linewidth]{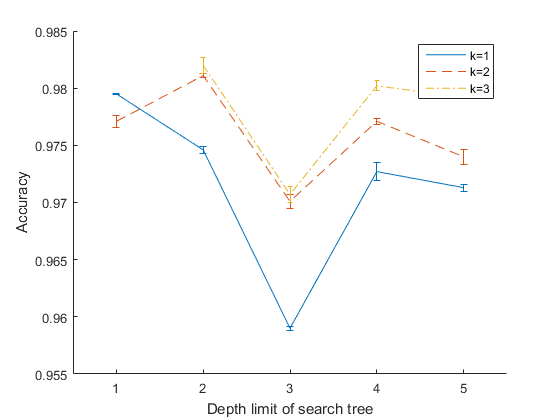}
\caption{3D shape classification accuracy on training data of 10 classes at different beam width $\textit{K}$.}
\label{fig:train1}
\end{figure}

\begin{figure}[!h]
\centering
\includegraphics[width=0.7\linewidth]{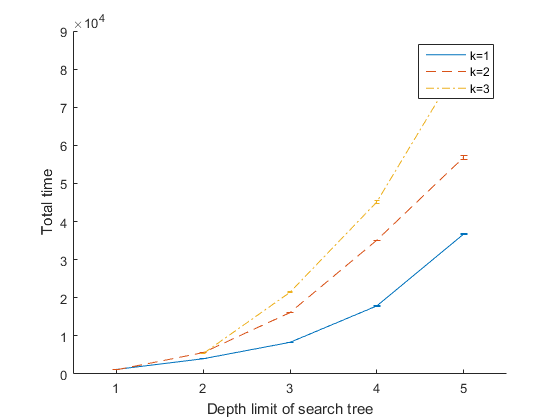}
\caption{Total search time for 10 classes dataset at different beam width $\textit{K}$.}
\label{fig:time1}
\end{figure}

\begin{figure}[!h]
\centering
\includegraphics[width=0.7\linewidth]{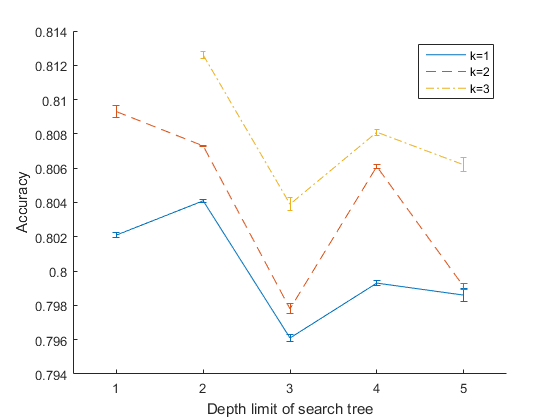}
\caption{3D shape classification accuracy on testing data of 40 classes at different beam width $\textit{K}$.}
\label{fig:test2}
\end{figure}

\begin{figure}[!h]
\centering
\includegraphics[width=0.7\linewidth]{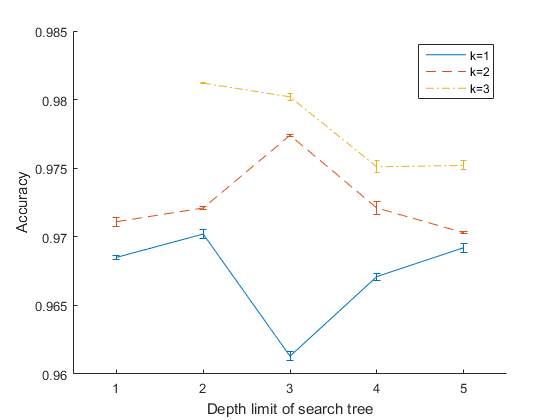}
\caption{3D shape classification accuracy on training data of 40 classes at different beam width $\textit{K}$.}
\label{fig:train2}
\end{figure}

\begin{figure}[!h]
\centering
\includegraphics[width=0.7\linewidth]{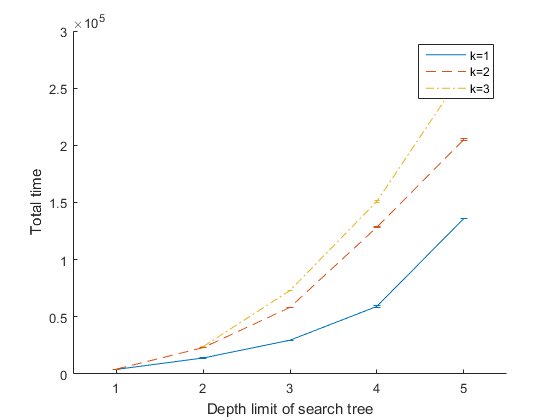}
\caption{Total search time for 40 classes dataset at different beam width $\textit{K}$.}
\label{fig:time2}
\end{figure}

We test how our performance varies for different depth limits $D = 1, 2, 3, 4, 5$, and beam widths $K = 1, 2, 3$. The training and testing accuracies, as well as the total beam-search runtime are presented in Figures \ref{fig:test1}, \ref{fig:train1}, \ref{fig:time1}, \ref{fig:test2}, \ref{fig:train2}, \ref{fig:time2}.

In the experiments, we observe that when considering the third type of action which adds a new max-pooling layer, this particular action is never selected by the beam search. This is in part due to the fact that adding a pooling layer results in re-initializing subsequent fully-connected layer.
In turn, this reduces the effectiveness of already learned parameters. Because of this, we actually do not consider the action of adding a pooling layer in our specification of the beam search.  

We compare our approach with two other approaches  3D ShapeNets \cite{3dshapenets} and DeepPano \cite{deeppano} in Table \ref{tab:accuracy}. As can be seen, on ModelNet10 and  ModelNet40,  our accuracy is by 3.63\% better than DeepPano in the 40-class experiment, and by 2.55\% better in the 10-class experiment.

\begin{table}[!h]
\renewcommand{\arraystretch}{1.5}
\begin{center} 
\begin{tabular}{|c|p{2cm}|p{2cm}|}
\hline
Algorithm & ModelNet40  Classification & ModelNet10  Classification\\
\hline
Ours & 81.26\% & 88.00\% \\
\hline
DeepPano \cite{deeppano} & 77.63\% & 85.45\% \\
\hline
3DShapeNets \cite{3dshapenets} & 77\% & 83.5\% \\
\hline
\end{tabular}
\end{center}

\caption{Comparison of our classification accuracy(\%) with state-of-the-art on the ModelNet10 and ModelNet40 datasets} \label{tab:accuracy}
\end{table}

We observe that our model produced by the beam search has much fewer parameters than the network used in \cite{3dshapenets}. Their model consists of three convolutional and two fully-connected learned layers. Their first layer has 48 filters of size 6;  the second layer has 160 filters of size 5; the third layer has 512 filters of size 4; the fourth layer is a fully connected RBM with 1200 hidden units; and the fifth and final layer is a fully-connected layer of size $C$, which is the number of classes. Our best found model consists of three convolutional layers and one fully-connected layer. Our first layer has 16 filters of size 6 and stride2; the second layer has 64 filters of size 5 and stride 2; the third layer has 64 filters of size 5 and stride 2; the last fully-connected layer has $C$ hidden units. Our model has about 80K/12M = 0.6\% parameters of theirs. 

The recent literature also presents two works on 3D shape classification:  VoxNet \cite{voxnet} and MVCNN \cite{mvcnn}, obtaining higher classification accuracies (90.1\%, 83\%) on ModelNet40  than ours. However, a direct comparison with these approaches  is not suitable.  VoxNet uses a training process that takes around 12 hours, while our individual training time for the best found model is less than 5 hours. MVCNN is based on the 2D information viewed from different angles around 3D shape, so it is inherently a 2D CNN approach but not related to 3D CNN. In addition, they also use the large collection of 2D images from ImageNet containing millions of images belonging to the same set of classes as the object categories presented in ModelNet40, to help their training process, while our work's only dataset is ModelNet40. So based on these reasons, we believe it is not suitable for our experimental results to be compared to theirs.

\section{Conclusion}
We have presented a new deep model pursuit approach  for 3D shape classification. Our learning uses a beam search, which explores the search space of various candidate CNN architectures toward achieving maximal classification accuracy. The search tree is efficiently built using a training classification accuracy based heuristic function, as well as knowledge transfer to efficiently estimate parameters of new candidate models. Our experiments demonstrate that our approach outperforms the state of  the art on the popular ModelNet10 and ModelNet40 3D shape datasets  by  3\% . Our approach also successfully reduces the total number of parameters  by 99.4\%. As our approach could be easily applied to other problems requiring robust deep learning on small training datasets.

\section*{Acknowledgment}
This research has been supported in part by National Science Foundation under grants IIS-1302700 and IOS-1340112.





\bibliographystyle{IEEEtran}
\bibliography{IEEEabrv,mybib}
%
%
%

\end{document}